\documentclass[11pt,a4paper]{article}
\usepackage[hyperref]{acl2021}
\usepackage{times}
\usepackage{latexsym}

\usepackage{microtype}
\usepackage{mathtools}
\usepackage{booktabs,graphicx}
\usepackage{caption,subcaption}

\usepackage{tikz}
\newcommand{\recipe}[2]{
\tikzstyle{mybox} = [draw=black, fill=white, thick,
    rectangle, rounded corners, inner sep=5pt, inner ysep=15pt]
\tikzstyle{fancytitle} =[fill=white, text=black, draw=black]

\begin{center}
\begin{tikzpicture}
\node [mybox] (box){%
    \begin{minipage}{0.95\columnwidth}
{#1}
    \end{minipage}
};
\node[fancytitle, right=10pt] at (box.north west) {#2};
\end{tikzpicture}%
\end{center}
}

\aclfinalcopy 
\def\aclpaperid{600} 

\title{As Good as New. How to Successfully Recycle English GPT-2 \\ to Make Models for Other Languages}

\author{Wietse de Vries \\
  University of Groningen \\
  The Netherlands \\
  \texttt{wietse.de.vries@rug.nl} \\\And
  Malvina Nissim \\
  University of Groningen \\
  The Netherlands \\
  \texttt{m.nissim@rug.nl} \\}

\date{}

\begin{document}
\maketitle
\begin{abstract}
Large generative language models have been very successful for English, but other languages lag behind, in part due to data and computational limitations.
We propose a method that may overcome these problems by adapting existing pre-trained  models to new languages.
Specifically, we describe the adaptation of English GPT-2 to Italian and Dutch by retraining lexical embeddings without tuning the Transformer layers.
As a result, we obtain lexical embeddings for Italian and Dutch that are aligned with the original English lexical embeddings.
Additionally, we scale up complexity by transforming relearned lexical embeddings of GPT-2 small to the GPT-2 medium embedding space.
This method minimises the amount of training and prevents losing information during adaptation that was learned by GPT-2.
English GPT-2 models with relearned lexical embeddings can generate realistic sentences in Italian and Dutch.
Though on average these sentences are still identifiable as artificial by humans, they are assessed on par with sentences generated by a GPT-2 model fully trained from scratch.
\end{abstract}

\section{Introduction}

Large pre-trained language models have brought unprecedented progress in NLP, but also concerns regarding the excessive computing power needed to train them \cite{strubell_energy_2019}.
Limited access to large amounts of computational resources, as well as  environmental considerations, curb possibilities for less-resourced and less-researched languages.
Additionally, models like GPT-2 \citep{radford_language_2019} are trained on amounts of data that are not available for most languages.
As a result of these limitations, language models are commonly trained for English, whereas reproductions in other languages may underperform or not exist.

That language models can benefit from information in other languages has been demonstrated by the effectiveness of multilingual BERT (mBERT) and XLM-RoBERTa \citep{conneau_unsupervised_2020}.
However, for downstream tasks mBERT has been shown to be outperformed by monolingual models for higher resource languages whereas lower resource languages can still achieve better results without pre-trained language models \citep{nozza_what_2020,wu_are_2020}.

Rather than pursuing a multilingual direction, we aim at exploiting existing language models and language similarities to create models for new languages. Specifically, 
we develop a multi-step procedure for adapting English GPT-2 \citep{radford_language_2019} to Italian and Dutch.
Dutch is genetically closely related to English, both being West-Germanic languages, while Italian is a more distant Romance language from the same Indo-European language family \citep{eberhard_ethnologue_2020}.
It is however worth noticing that at sentence level English and Italian tend to have the same word order (SVO), while Dutch is SVO in main clauses, but SOV in subordinate ones; at noun phrase level, English and Dutch share constituent order (for example adjective-noun) while Italian is different (mostly noun-adjective).
A GPT-2 based model has previously been trained from scratch for Italian \citep{de_mattei_geppetto_2020}. We can thus compare sentences generated by this model with sentences generated by our adapted model.
For Dutch, no other GPT-2 based models exist, but similar BERT-based models have been trained from scratch \citep{de_vries_bertje_2019, delobelle_robbert_2020}.

\paragraph{Procedure Overview and Contributions} When training a new language model, weights of an existing pre-trained model for another language can be used for initialisation.
The first step in our training procedure is to only retrain the lexical embeddings of the GPT-2 \textit{small} model, without touching the Transformer layers.
We show that retrained lexical embeddings are well aligned with the English vocabulary and that GPT-2 is capable of generating realistic text in Italian and Dutch after this step.
Next, we demonstrate that the lexical embeddings of larger GPT-2 models can be approximated by transforming the \textit{small} lexical embeddings to the GPT-2 \textit{medium} lexical embedding space.
The least-squares regression method is the most effective transformation method for this scaling procedure.
Human judgements show that generated sentences are often realistic, but become even more consistently so after additional finetuning of the Transformer layers.
This improvement is stronger for Dutch than for Italian.

The steps in our pipeline yield GPT-2 based language models for Italian and Dutch which are made available on the Hugging Face model hub\footnote{\url{https://huggingface.co/GroNLP}};
the source code is available on Github\footnote{\url{https://github.com/wietsedv/gpt2-recycle}}.
On the last page, we also include a `recipe' for creating GPT-2 models for new languages.

\section{Background}

Previous and current research relevant for the present work is found in the more general field of transfer learning, with a specific focus on language transfer.
We also discuss how our approach of translating lexical layers in different model sizes relates to work on aligning word embeddings.

\subsection{Language transfer}
Transfer learning can be an effective strategy to adapt models to lower-resource languages by initially training a model for a source language and then further training (parts of) the model for a target language.
It has been successfully used to create machine translation models with little parallel data \citep{zoph_transfer_2016} as well as other classic NLP tasks \citep{lin_choosing_2019}.

In machine translation a model can be adapted by initially training it for a high-resource language pair after which the model should be partially retrained for a low-resource language \citep{zoph_transfer_2016, nguyen_transfer_2017, kocmi_trivial_2018}.
Retraining a randomly initialised lexical layer while freezing the rest of the model is an effective method to adapt a model to a new language, and dictionary based initialisation is not required to get the best performance \citep{zoph_transfer_2016}.
\citet{artetxe_cross-lingual_2020} show that a monolingual BERT model can be adapted from a source language to a different target language by retraining the lexical layer for the target language while freezing the Transformer layers in the model.
Zero shot adaptation for downstream tasks is  possible by finetuning the original source model with source language data and swapping lexical layers afterwards.
Lexical layer retraining approaches may be effective despite the presence of source and target language dissimilarities if a downstream task does not require perfect data.
However, these methods have not been applied yet to generative language models where dissimilarities can cause clear syntactic and lexical errors.

Language similarity plays a role in the effectiveness of transfer learning for language models. 
For instance, in machine translation French is a better parent model for Spanish than German \citep{zoph_transfer_2016}.
Word order differences between languages can negatively influence transfer performance, and \citet{kim_effective_2019} show that randomly swapping words in the source language, which forces the model to rely less on consistent word order, can improve performance in the target language.
Overall, genetic similarity between source and target languages can play a role, but  \citet{lin_choosing_2019} have shown that in practice the geographic distances between countries of origin, syntactic similarity and subword overlap are better predictors of transfer performance for machine learning, part-of-speech tagging, dependency parsing and entity linking.


\subsection{Aligning word embeddings}
Alignment of lexical embeddings, for example for multiple languages, is most prominently done with mapping-based approaches \citep{ruder_survey_2019}.
Typically, a function is determined that transforms one vector space to another based on a seed lexicon.
This lexicon is a dictionary of anchor points that should result close together after transformation.

An influential method for learning a lexical embedding mapping is the least-squares linear transformation method by \citet{mikolov_exploiting_2013}. 
They observe that words and their translations in other languages show similar constellations of related words after such a transformation. 
An alternative method that is generally considered to be an improvement \citep{ruder_survey_2019} is the orthogonal procrustes solution.
This method adds the constraint  that the transformation matrix must be orthogonal.
In practice this means that the transformation only contains rotations and reflections and no scaling and translation.
This constraint enables length normalisation \citep{xing_normalized_2015} and ensures monolingual invariance \citep{artetxe_learning_2016}.

Mapping-based approaches rely on isomorphism, which means that a one-to-one token mapping between source and target lexical embedding spaces should be possible.
This assumption is used for bilingual lexicon induction after alignment \citep{conneau_word_2018}.
However, 
the isomorphism assumption 
highly depends on language similarity and (amount of) training data \citep{sogaard_limitations_2018}.
Some more complex alignment methods like RCLS \citep{joulin_loss_2018} optimise for dictionary translation performance, which assumes isomorphism, but simpler methods like the orthogonal procrustes solution are more effective for downstream tasks like natural language inference \citep{glavas_how_2019}.
\citet{mohiuddin_lnmap_2020} propose a solution to the isomorphism problem by learning a new shared embedding space with an auto-encoding neural model instead of trying to fit the embeddings of one language in the space of another language.

\section{Resources}

\paragraph{Models}
The models that we train are based on the pre-trained GPT-2 language models \citep{radford_language_2019}.
GPT-2 is an auto-regressive Transformer-decoder based language model for English and comes in four sizes: small (12 layers), medium (24 layers), large (36 layers) and extra large (48 layers).
Our experiments use the small (\texttt{sml}) and medium (\texttt{med}) model sizes.

\paragraph{Pre-training data}
The GPT-2 models are (further) pre-trained with Italian (\texttt{ita}) and Dutch (\texttt{nld}) data.
The Italian pre-training data is the same dataset that was used to train  the Italian GPT-2 small language model GepPpeTto \citep{de_mattei_geppetto_2020}.
This dataset is a combination of Wikipedia data (2.8GB) and web texts from the ItWaC corpus (11GB; \citealt{baroni_wacky_2009}).
Dutch data consists of a combination of Wikipedia (2.0GB), newspaper articles (2.9GB; \citealt{ordelman_roeland_jf_twnc_2007}), books (6.5GB) and articles from various Dutch news websites (2.1GB).
Documents are filtered to only contain Dutch texts using the Wikipedia-trained fastText language identifier \citep{joulin_bag_2017}, and are deduplicated based on exact sentence matches.
The final Dutch pre-training data contains 13GB of plain text, of which 5\% is reserved as development data.

\paragraph{Evaluation data}
The Italian models are tested using the same three  corpora that were used to evaluate  GePpeTto \citep{de_mattei_geppetto_2020}: Wikipedia, ItWaC, EUR-Lex (laws), newspapers and blog posts.
A 5\% subset of this  data is used for development.
For perplexity evaluation, the Dutch 500 million word, 22-genre SoNaR corpus is used \citep{oostdijk_construction_2013}.
The smaller 1 million word SoNaR-1 subcorpus is used as development data.

\paragraph{Tokenisation}
The datasets are tokenised using byte-pair-encoding (BPE).
For better comparison, the Italian vocabulary is taken from the GePpeTto model \citep{de_mattei_geppetto_2020}.
The Dutch BPE vocabulary is based on the full pre-training data and it has been ensured that every character that is used in the Dutch language is present as a single character token in the vocabulary.
A large vocabulary size is beneficial because words are  less often split in separate tokens, but vocabularies that are too large will have low token coverage for uncommon tokens.

\paragraph{Computation}
Training a model like GPT-2 is a computationally expensive task that requires access to costly hardware for long training times.
All models discussed in this paper are trained with eight parallel NVIDIA V100 32GB GPUs on the Peregrine high performance computing cluster at the University of Groningen.
For efficient implementation of the models, we use PyTorch (1.6.0; \citealt{paszke_pytorch_2019}), PyTorch Lightning (0.9.0; \citealt{falcon_pytorch_2019}) and Transformers (3.0.2; \citealt{wolf_huggingfaces_2020}).
We implement four strategies to decrease general training time.
First, the models are trained with 16-bit automatic mixed-precision training \citep{micikevicius_mixed_2018}. This decreases training time with a factor of two to three times.
Second, we split each document in windows of 128 instead of 1024 tokens when we only train the lexical embeddings.
Third, we minimise padding by using bucketed random sampling which means that sequences within minibatches have roughly the same length.
Finally, we use maximum batch sizes that fit into GPU memory and use gradient accumulation in order to do backpropagation only for every 2000 examples.
The models are trained with the Adam optimiser \citep{kingma_adam_2017} and initial learning rates are chosen based on the steepest loss slope with gradually increasing learning rates \citep{smith_cyclical_2017}.
The learning rate is reduced by 10\% on when training loss reaches a plateau.
More implementation details are given in the git repository.

\section{Cross-language Transfer}
\label{sec:lang}

We adapt GPT-2 for Italian and Dutch with minimal random initialisation.
The lexical embeddings in GPT-2 are trained with an English BPE vocabulary. Therefore, they are  not usable for the new languages and the lexical embedding layer has to be randomly initialised for the target vocabulary.
This lexical embedding layer is used both as the first and the last layer of GPT-2 (tied weights).
Relearning lexical embeddings with frozen Transformer layers prevents catastrophic forgetting in the Transformer layers when the embeddings are still random.


\paragraph{Relearning lexical embeddings}
Relearning lexical embeddings is nearly as computationally expensive as fully training the model, because back-propagation has to be done through the full model in order to update the lexical embeddings in the first layer of the model.
However, loss values stabilize after only one to two epochs with lexical embedding relearning whereas full model training takes more training time.
We retrain the lexical embeddings for the \texttt{sml} and \texttt{med} model for Italian and Dutch by training until loss on the validation data stops decreasing.
When we retrain the \texttt{sml} model, the perplexities on our Italian and Dutch test data become 44.2 and 48.9 respectively.
These perplexity scores show that the \texttt{sml} model can predict Dutch and Italian tokens reasonably well without having retrained the Transformer layers.
Therefore, the English Transformer layers are at least partially language-independent and our relearning method automatically aligns lexical embeddings to the embedding space of the English model.
However, if we retrain the \texttt{med} lexical layer for Italian and Dutch with the same method, test data perplexities are 81.2 and 185.0.
These unsatisfactory \texttt{med} perplexities could be due to stopping training too early or to arriving at a suboptimal local optimum.
Training for a longer time or trying different random initialisations defeats the purpose of minimising computational requirements.
A more efficient method that uses the already learned \texttt{sml} embeddings is described in Section~\ref{sec:complexity}.

\begin{table}
\centering
\begin{tabular}{l | l l}
\toprule
\textbf{English} & \textbf{Italian} & \textbf{Dutch} \\
\midrule
while		& mentre & terwijl \\
genes	 & geni	& genen \\
clothes		& vestiti & kleren \\
musicians	 & composi[...] & artiesten \\
permitted & ammessa 	& toegelaten \\
Finally	& infine	& Eindelijk \\
satisfied	& soddisfatto	& tevreden \\
\midrule
\textit{Accuracy:} & \multicolumn{1}{c}{85\%} & \multicolumn{1}{c}{89\%} \\
\bottomrule
\end{tabular}
\caption{\label{tab:wte:alignment} Alignment of closest tokens in the lexical embeddings of \texttt{sml\textsubscript{rle}} for Italian and Dutch. Accuracy scores are based on a manual evaluation by the authors of 200 random aligned tokens. Semantically correct subword matches are included.}
\end{table}

\begin{table*}
\centering
\begin{small}
\begin{tabular}{p{0.98\columnwidth} | p{0.98\columnwidth}}

\toprule
\textbf{Italian} & \textbf{Literal English translation} \\
\midrule
La prima parte del film venne \textit{distribuito} in Giappone con l'aggiunta della colonna sonora. & The first part of the film was \textit{distributed} in Japan with the addition of the soundtrack.\\\midrule
L'unico motivo \textit{di la} mia insoddisfazione fu il fatto che l'inizio della sua attività [\ldots]& The only reason \textit{of the} my unsatisfaction was the fact that the beginning of-the his/her activity [\ldots]\\\midrule
Il suo nome deriva da un vocabolo arabo. & The his/her name derives from a word Arabic.\\

\toprule
\textbf{Dutch} & \textbf{Literal English translation} \\
\midrule
In een artikel in de Journal of Economicologie (1998), \textit{The New York Times schrijft}: & In an article in the Journal of Economicology (1998), \textit{The New York Times writes}:\\\midrule
Ik kan me niet voorstellen dat mensen van mijn generatie \textit{zijn zo boos op mij te wachten}. & I can me not imagine that people of my generation \textit{are so mad at me to wait}.\\\midrule
Ik heb niets gedaan om mijn moeder te helpen. & I have nothing done to my mother to help.\\

\bottomrule
\end{tabular}
\end{small}

\caption{\label{tab:wte:examples} A selection of generated sentences by the \texttt{sml} model with Italian and Dutch lexical embeddings. Phrases in italics are ungrammatical in the target language.}
\end{table*}

\paragraph{Vocabulary alignment}
The lexical embeddings of both the original English tokens as well as the relearned Italian and Dutch lexical embeddings can be considered to inhabit the same embedding space because the lexical embeddings of all three languages are tuned to minimise loss with the exact same Transformer layers.
Therefore, tokens with similar meaning in different languages should be close to each other if the lexical embeddings are properly trained.
Table~\ref{tab:wte:alignment} shows the closest Italian and Dutch tokens of a random sample of English tokens.
These alignments show that the optimal lexical embeddings for both Italian and Dutch are often literal translations of English tokens.
Thanks to similarity of context-dependent structures like syntax in these three languages, the English model can be adapted to Italian and Dutch.
Based on this small sample, Dutch to English alignment seems to be slightly more accurate than Italian to English, but a more thorough study would be required to evaluate the actual relation between genetic similarity and alignment potential through this method.

\paragraph{Text generation}
Table~\ref{tab:wte:examples} shows some examples of unconditioned text generation of the English \texttt{sml} with relearned lexical embeddings for Italian and Dutch.
These examples show that the model can generate proper Italian and Dutch sentences, although it sometimes uses English word order where the correct word order differs in Dutch, or ignores grammatical gender agreement in Italian defaulting to the singular masculine, or doesn't always produce correctly Italian prepositional articles (``di la'', en: \textit{of the} vs ``della'', en: \textit{of-the}). Phrases in italics in  Table~\ref{tab:wte:examples} highlight such mistakes.

The literal English translations, however, 
show that the models can generate proper Italian and Dutch grammar that differs from English.
Italian and Dutch lexical embeddings are not only aligned with equivalent English tokens, but unexpected correct syntax shows that the grammatical functions of words have also been adapted. For example, in Italian the noun-adjective order is opposite to English and realised correctly; also, the use of the definite article in front of a possessive pronoun is correctly introduced, while ungrammatical in English.

This shows that the relatively low-dimensional context-independent lexical embeddings in GPT-2 contain syntactic features of the tokens in addition to semantics, and confirms previous findings of high information density in the lexical layer of language models \cite{de_vries_whats_2020}.
Therefore, language adaptation can be to some extent effective by adapting the lexical embedding layer without retraining Transformer layers at all.

\section{Scaling up Complexity}
\label{sec:complexity}

Replacing the original lexical embeddings with lexical embeddings from a different target language seems an effective way to initialise full model transfer to that target language.
However, relearning the lexical embeddings of a new vocabulary requires full forward and backward propagation through the whole model.
Therefore, this becomes an increasingly more expensive task for larger model sizes.
When multiple model sizes need to be transferred to a new language, the lexical embeddings do not need to be retrained from scratch.
Instead, vocabulary alignment between the source and target languages for the smaller model could be used to initialise the embeddings for a larger model.

After relearning the lexical embeddings of the \texttt{sml} model for Italian and Dutch, we observed that tokens with similar meaning in different languages are close to each other in the embedding space.
This alignment effect should also be present in properly trained lexical embeddings of larger models.
Given that we have at our disposal known embeddings for all 50K English tokens for every model size, we can use these data points to transform model size \texttt{sml} to larger model size \texttt{med}.

Regardless of architecture, embeddings are only considered to be alignable if they are trained under identical conditions with the same type and amount of data \citep{levy_improving_2015, ruder_survey_2019}.
Our goal differs from previous alignment efforts since instead of aligning languages, we align separately trained embeddings for different model sizes, trained on the same data with identical and fully parallel vocabularies in English. The embeddings differ in dimensionality (768d for \texttt{sml}, 1024d for \texttt{med}) and the different model sizes may influence the amount and density of information in the lexical embeddings.

\begin{table*}[ht]
\centering
\begin{tabular}{l | c c | c c c }
\toprule
  & \multicolumn{2}{c|}{Italian} & \multicolumn{3}{c}{Dutch} \\
 \textbf{Model}         &   \textbf{Int@1k} & \textbf{PPL}  &   \textbf{Int@1k} & \textbf{PPL} & \textbf{PPL (1 epoch)} \\
\midrule
  \texttt{med\textsubscript{rle}} (1 epoch) & 0.38 & - & 185.02                         & - & - \\
\midrule
 \texttt{sml\textsubscript{rle}} $\xrightarrow{proc}$ \texttt{med}   & \textbf{0.61} & 8.12 $\times 10^{12}$ & \textbf{0.61} & 5.02 $\times 10^{12}$ & 52.69 \\
 \texttt{sml\textsubscript{rle}} $\xrightarrow{lstsq}$ \texttt{med}  & 0.56 & \textbf{364.06} & 0.56 & \textbf{293.61}                & \textbf{47.57} \\
 \texttt{sml\textsubscript{rle}} $\xrightarrow{1-nn}$ \texttt{med}   & 0.37 & 2,764.19 & 0.36 & 1,101.59                       & 50.25 \\
 \texttt{sml\textsubscript{rle}} $\xrightarrow{10-nn}$ \texttt{med}  & 0.37 & 20,715.80 & 0.35 & 11,871.66                      & 56.88 \\
\bottomrule
\end{tabular}%
\caption{\label{tab:trans:med} Scores for different transformation methods. Int@1K are the average 1k nearest English neighbours intersection (int) fractions between \texttt{sml} and transformed \texttt{med} embeddings. \textit{PPL} is the perplexity on the test sets for Italian and Dutch. \textit{PPL (1 epoch)} indicates the perplexity after one epoch of training, which is low if the transformed embeddings were close to a good local optimum.}
\end{table*}

\subsection{Transformation methods}
The 50K parallel English tokens can be used to find an optimal transformation between lexical embeddings of different model sizes.
The completeness of this mapping due to shared vocabularies between models eliminates the need to use complex solutions like refinement or bootstrapping the lexicon \citep{artetxe_unsupervised_2018}.
We compare three simple supervised alignment methods for transformation from source space \texttt{sml} to target space \texttt{med}.

\paragraph{Regression (lstsq)}
A classic approach for mapping lexical embeddings is mean-squared-error minimising linear regression with the least-squares method \citep{ruder_survey_2019, mikolov_exploiting_2013}.
This method learns a transformation matrix $\boldsymbol{W}$ that minimises the Euclidean distance between source and target embeddings.
The optimal matrix is approximated with stochastic gradient descent, and therefore this is not an exact solution.

\paragraph{Orthogonal Procrustes (proc)}
More recent alignment approaches constrain the transformation $\boldsymbol{W}$ to be an orthogonal matrix \citep{ruder_survey_2019, artetxe_learning_2016}.
This constraint enables using the exact solution for the orthogonal Procrustes problem \citep{xing_normalized_2015}.
The exact solution only rotates and reflects data points to be as close as possible to the target space without any scaling or translation, preserving monolingual invariance in the source embeddings \citep{artetxe_learning_2016}.

\paragraph{Weighted K-Nearest Neighbours (knn)}
Unlike typical alignment approaches, we have a complete set of parallel data points in the source and target spaces (English).
The unknown target language tokens can be approximated by taking the $\boldsymbol{K}$ nearest English tokens in the source \texttt{sml} embedding space and using the distance-weighted sum of these tokens in the target \texttt{med} embedding space.

\subsection{Results after transformation}

Table~\ref{tab:trans:med} shows \texttt{med} embedding similarity with source \texttt{sml} embeddings and perplexities on test data with transformed embeddings, as well as the transformed embeddings with one additional epoch of training.
Results are consistent for Italian and Dutch.
Nearest English neighbours are best preserved with the Orthogonal Procrustes method.
However, the perplexity scores are extremely high for this method.
The perplexity scores with the different methods vary in complete orders of magnitude.
Based on this, the least-squares regression method outperforms the other methods.
After one epoch of additional training with transformation initialised embeddings, the lstsq method still outperforms the other methods.
It even outperforms the \texttt{sml} model with fully tuned lexical embeddings.

\begin{table}
\centering
  \resizebox{7.4cm}{!}{%
\begin{tabular}{l | c r }
\toprule
\textbf{Model} & \multicolumn{2}{c}{\textbf{PPL}}  \\
               & \texttt{ita} & \texttt{nld} \\
\midrule
 \texttt{sml\textsubscript{rle}}  & 44.19 & 48.85 \\
 \texttt{sml\textsubscript{rle} + finetuning} & \textbf{42.45} & \textbf{39.59} \\
 \texttt{sml\textsubscript{full}}* & 193.15 & 219.34 \\
\midrule
 \texttt{med\textsubscript{rle} + finetuning}  & 42.51  & 44.68 \\
\midrule
 \texttt{GePpeTto (sml)}  & 106.84 & -  \\
\bottomrule
\end{tabular}%
}
\caption{\label{tab:full} Perplexities of the concatenated test data for the final models. The \texttt{med\textsubscript{rle}} model is in practice the \texttt{sml\textsubscript{rle}} $\xrightarrow{lstsq}$ \texttt{med} model. * The \texttt{sml\textsubscript{full}} model is trained for the equivalent amount of time as the \texttt{sml\textsubscript{rle} + finetuning} models, but with all layers unfrozen.}
\end{table}

\section{Full model finetuning}
\label{sec:full}

After obtaining lexical embeddings for Italian and Dutch to be plugged into the English GPT-2 models, the full models can be finetuned for the target language.
The best performing lexical embeddings will be used to train the \texttt{sml} and \texttt{med} Italian and Dutch models.
These are the lexical embeddings that are relearned from random initialisation for the \texttt{sml} model.
For the \texttt{med} model, the lstsq transformed \texttt{sml} embeddings with additional training are used (\texttt{sml\textsubscript{rle}} $\xrightarrow{lstsq}$ \texttt{med\textsubscript{+rle}}).

The relearned lexical embeddings reduce the risk of information loss while the model is adjusting to a new language.
Nevertheless, information can still be lost during training.
For instance for the \texttt{sml} Dutch model, validation loss increases with a learning rate of $10^{-4}$, but this does not happen with a lower learning rate of $10^{-5}$.

\section{Obtained models and evaluation}
\label{sec:eval}

For both Italian and Dutch, we evaluate three models: 
(i) the English \texttt{sml} model with relearned lexical embeddings; 
(ii) the \texttt{sml} model with additional finetuning to the target language; and 
(iii) the English \texttt{med} model with relearned lexical embeddings that were initialised by transforming \texttt{sml} embeddings with the least-squares method.
For Italian, we also include the GPT-2 small based GePpeTto model \citep{de_mattei_geppetto_2020}, which was trained from scratch.
This inclusion offers the opportunity of a direct comparison between a GPT-2 model trained from scratch and those obtained with our transfer approach.
We run both an automatic and a human-based evaluation.
For the former, we compare perplexity scores on unseen test data in different genres.
For the latter, we collect and compare judgements over generated and gold texts by native speakers of Italian and Dutch.

\subsection{Perplexity}
\label{sec:res:ppl}

Table~\ref{tab:full} shows perplexity scores on concatenated multi-genre test data based on a strided moving window perplexity calculation.\footnote{Window sizes are 128 tokens and strides are 64 tokens except for GePpeTto. GePpeTto was trained with at most 100 tokens, so its window size is 100 with a 50 token stride.}
Perplexities are calculated with Italian and Dutch vocabularies of 30K tokens.
These results show that perplexities are low when only relearning the lexical embeddings for both Italian and Dutch.
Further finetuning of the \texttt{sml} model seems to have the greatest effect for the Dutch language.
The \texttt{med} models with relearned lexical embeddings have lower perplexity than the equivalent \texttt{sml} models.
This shows that language transferability based on the lexical layer is not restricted to small model sizes.
Moreover, we see that our proposed method results in lower perplexity scores than regular full model finetuning of the English model.
The overall perplexity scores of Italian are closer to each other than the Dutch perplexities.
We also tested perplexities by the different genres that make up both the Italian and the Dutch datasets (see Figure~\ref{tab:res:ita:genres} and Figure~\ref{tab:res:nld:genres} for details), and observed that while perplexities vary greatly per genre, the model ranking per genre is consistent with the global scores.


\begin{table}[ht!]
\centering
\resizebox{7.7cm}{!}{%
\begin{tabular}{l | c c c }
\toprule
\textbf{Model} & \textbf{Social} & \textbf{News} & \textbf{Legal} \\
\midrule
 \texttt{sml\textsubscript{rle}} & 134.64 & 67.14 & 16.95 \\
 \texttt{sml\textsubscript{rle} + finetuning} & \textbf{118.19} & \textbf{55.63} & 15.36 \\
\midrule
 \texttt{med\textsubscript{rle}} & 123.64 & 59.18 & \textbf{14.95} \\
\midrule
 \texttt{GePpeTto\textsubscript{sml}} & 179.47 & 80.83 & 34.71 \\
\bottomrule
\end{tabular}%
}
\caption{\label{tab:res:ita:genres} Perplexities for different genres within the Italian test data. Rankings are consistent with Table~\ref{tab:full} except for the legal domain.}
\end{table}

\begin{table}[ht!]
\centering
\resizebox{7.7cm}{!}{%
\begin{tabular}{l | c c c }
\toprule
\textbf{Model} & \textbf{Proceedings} & \textbf{News} & \textbf{Legal} \\
\midrule
 \texttt{sml\textsubscript{rle}}        & 44.47 & 239.14 & 52.01 \\
 \texttt{sml\textsubscript{rle} + finetuning} & \textbf{36.35} & \textbf{171.83} & \textbf{42.92} \\
\midrule
 \texttt{med\textsubscript{rle}}        & 40.62 & 234.52 & 45.01 \\
\bottomrule
\end{tabular}%
}
\caption{\label{tab:res:nld:genres} Perplexities for some SoNaR genres in Dutch. Models rankings are consistent across genres.}
\end{table}

\subsection{Human Judgements}
\label{sec:human}

The perplexity scores give an indication on how well a language is represented by language models, but this does not reliably tell how good the model is in a generative setting. For this, we resort to human judgements.
Human assessments of generated texts are collected for the models that incorporate the crucial steps in our approach and achieve reasonable perplexity scores: the \texttt{sml} models with only relearned lexical embeddings, the finetuned \texttt{sml} models and the higher complexity \texttt{med} models with only relearned lexical embeddings based on transformed \texttt{sml} lexical embeddings.

Texts are assessed in isolation by means of a \textit{direct} evaluation \cite{novikova_rankme_2018}.\footnote{A direct evaluation is opposed to a comparative one, usually involving a ranking task \cite{novikova_rankme_2018,de_mattei_geppetto_2020}; this is left to future work.} 
Subjects are presented with texts on the screen, and are asked  whether the texts they see could have been written by a human. All subjects are pre-informed that some of the texts they will see are machine generated.
Rather than discrete answers, we obtain continuous evaluations by offering the possibility of clicking anywhere on a bar whose extremes are ``no'' to the left and ``yes'' to the right.

The evaluation interface is made with PsychoPy3 \citep{peirce_psychopy2_2019} and hosted with Pavlovia\footnote{\url{https://pavlovia.org}}.

\begin{figure*}[t!]
     \centering
     \begin{subfigure}[b]{\columnwidth}
         \centering
         \includegraphics[width=\textwidth]{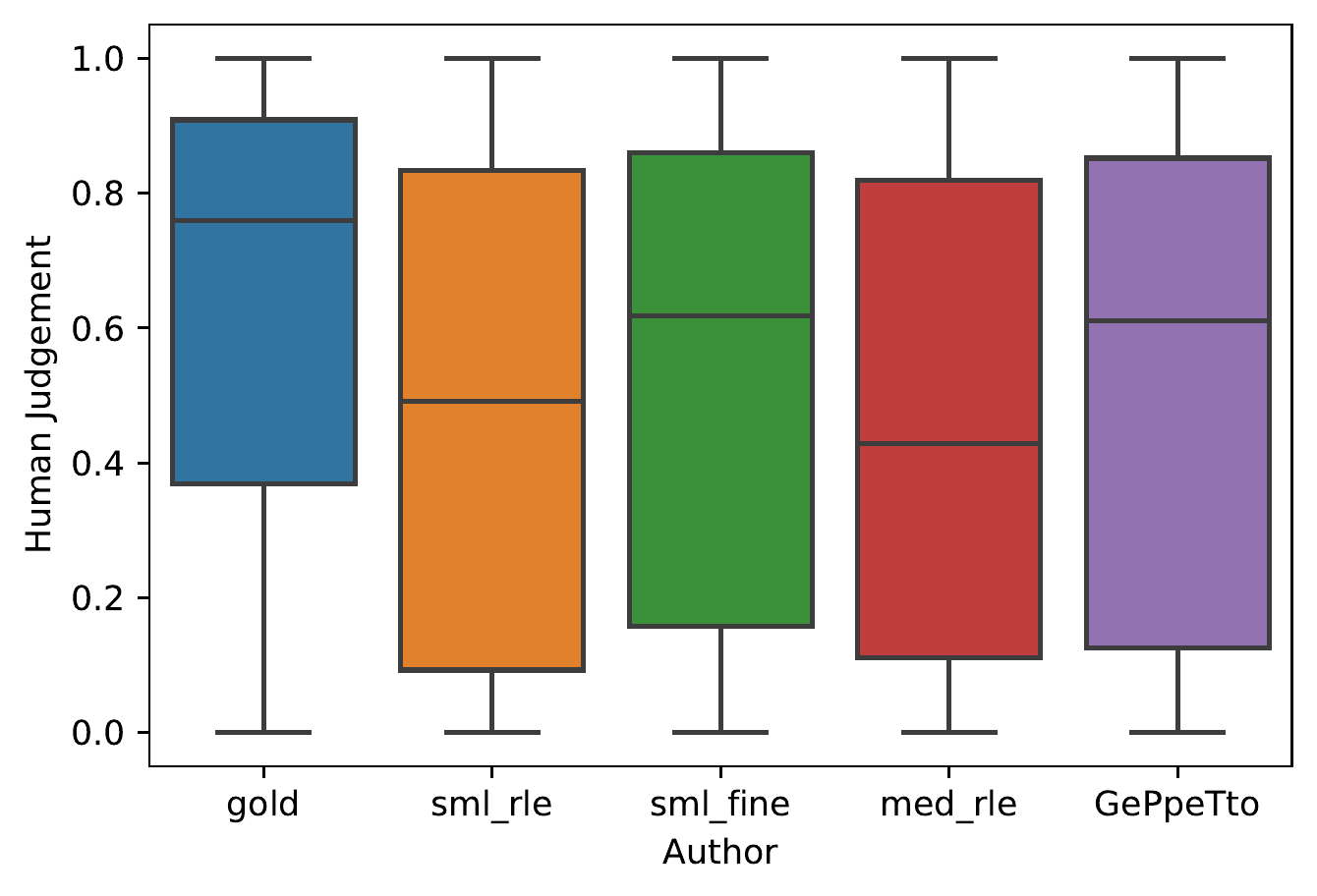}
         \caption{Human judgement scores for Italian texts.}
         \label{fig:human:ita}
     \end{subfigure}
     \begin{subfigure}[b]{\columnwidth}
         \centering
         \includegraphics[width=\textwidth]{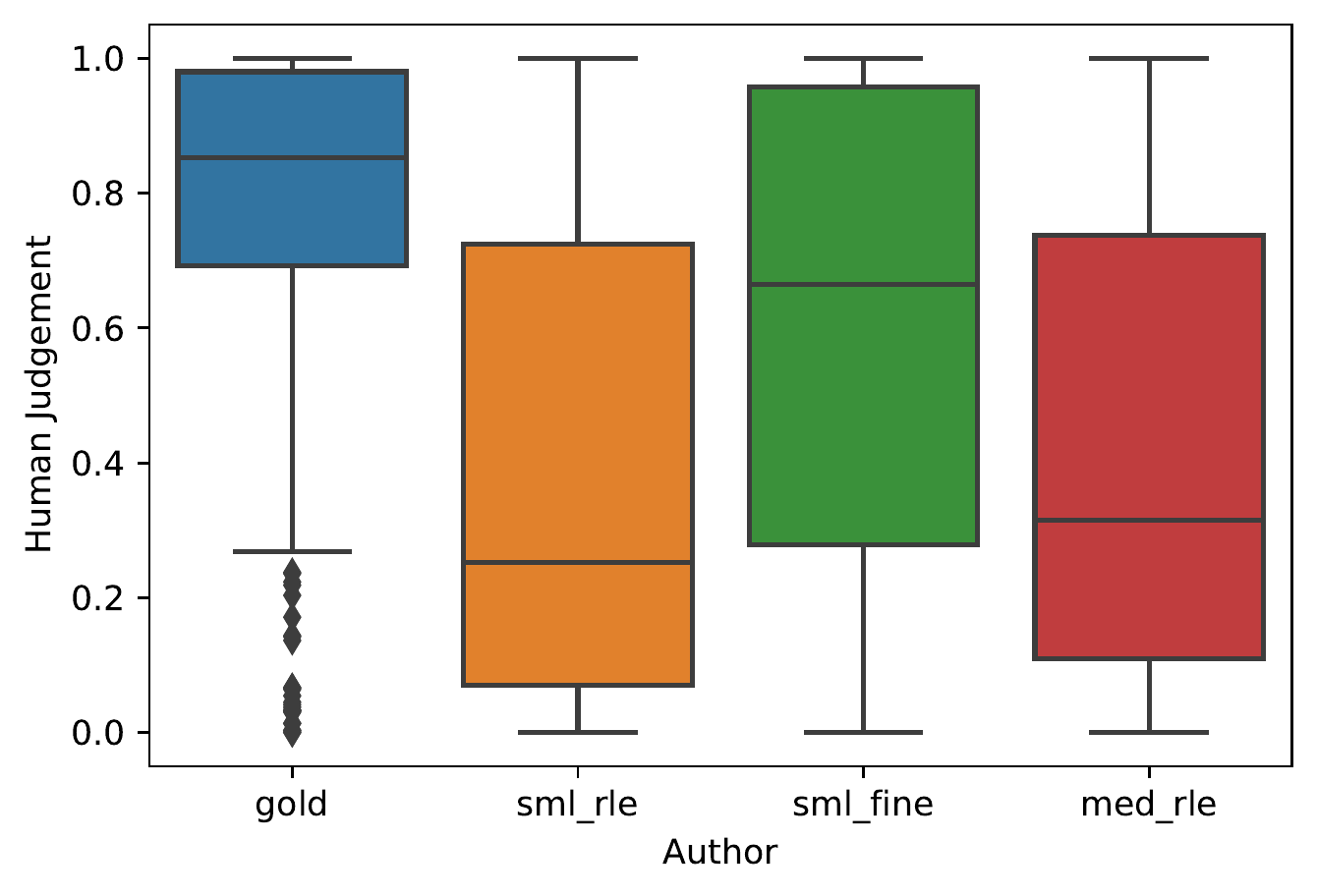}
         \caption{Human judgement scores for Dutch texts.}
         \label{fig:human:nld}
     \end{subfigure}
     \caption{Human judgement scores based on a continuous scale. Most judgements were close to 0 or 1.}
     \label{fig:human}
\end{figure*}

Italian models were evaluated by 24 participants (9~M, 15~F) with ages ranging from 26 to 63 with a median age of 46.
The Dutch models were evaluated by 15 participants (11~M, 4~F) with ages ranging from 23 to 36 with a median age of 27.

The three final models are evaluated for both languages; for Italian, we also add GePpeTto  \citep{de_mattei_geppetto_2020}.
Human written gold sentences were sampled from the test data as an additional condition.
For each of these 5 Italian and 4 Dutch conditions, 100 sentences are evaluated.
Each participant has evaluated 50 to 150 sentences and each sentence is evaluated by 3 to 5 participants. As a result, we obtain 1950 evaluations for 500 Italian texts and 1550 evaluations for 400 Dutch texts.

All artificial sentences are randomly generated without conditioning and with beam search (5 beams, with top 50 tokens or a summed probability of at least 90\%), and a temperature of 3.0.
Setting the temperature value $>$1 means decreasing the sampling probability of likely tokens, and therefore increases variation between generated samples.

Longer sentences have a higher chance to contain mistakes, so a model that generates longer sentences may have a disadvantage.
However, explicitly controlling sentence length is not possible nor desired since sentence length may also be an indication of model quality.
For both languages the randomly sampled gold sentences have more long sentences than the models, but the \texttt{sml} model with  finetuning also sometimes generates longer sentences.
We filter out sentences longer than 30 tokens to decrease sentence length effects on judgements.
The remaining Italian sentences have median lengths of 18 or 22 words and the Dutch ones 16 or 17 words for the different conditions.


Figure~\ref{fig:human} shows the distributions of human judgements per condition.
Variance seems to be high due to the non-normally distributed scores as relatively many scores are close to zero, half or one.
The model differences appear stronger for Dutch than Italian, but for both languages the subjects have given high scores to gold sentences.
This is expected and indicates that the participants are able to correctly judge real human texts.
Of the three trained models, the small model with additional finetuning achieves the highest scores.

For the Italian model comparison we use a linear mixed-effects model but with only author as fixed effect and random intercepts for participants and sentences.
There is no significant effect for sentence length.
The judgements on gold texts are significantly higher than all model judgements ($p < 0.005$) except for  \texttt{sml\textsubscript{fine}}.
However, \texttt{sml\textsubscript{fine}}  is not significantly better than GePpeTto nor the \texttt{sml\textsubscript{rle}} and \texttt{med\textsubscript{rle}} models ($p > 0.05$).

For Dutch we use a linear mixed-effects model with fixed effects for author and sentence length (in number of words) and random intercepts for participants and sentences.
Sentence length has a significant negative effect ($p < 0.001$).
All artificial authors score significantly lower than gold ($p < 0.001$).
As for Italian, the \texttt{sml\textsubscript{fine}} model appears the best  model, but in this case the  judgement scores are significantly higher than for the other two models ($p < 0.001$).
The \texttt{sml\textsubscript{rle}} and \texttt{med\textsubscript{rle}} models do not differ significantly from each other.

The human judgements show consistent results across the languages, but differences between Dutch judgements are stronger than for Italian.
This seems to mirror the smaller perplexity differences for Italian than for Dutch.
Whether demographic or cultural differences also play a role in this difference will need to be further investigated.

In sum, we see that the English GPT-2 models with relearned lexical embeddings are recognisable as artificial, whereas this problem is attenuated after additional finetuning.
The \texttt{sml} model with additional finetuning performs at least as well as the GePpeTto model that was trained from scratch.

\section{Conclusion}
We have described methods to adapt GPT-2 to genetically related languages and to increase model complexity.
Retraining lexical embeddings forces the model to learn representations that are aligned between English and the target language.
GPT-2 is able to generate realistic text in another language, but human judgements reveal that additional finetuning of the full model is needed to generate realistic sentences more consistently.
Relearned lexical embeddings show signs of syntactic adaptation to the new language, though not fully consistently.

Dutch is genetically closer to English than Italian, but our results do not prove that this method works better for Dutch.
Future research on the relation between degrees and types of language similarity and transferability of models will enable more effective monolingual transfer, and possibly training better multilingual models by selecting optimal clusters of languages.
This kind of work offers a privileged perspective into the information learned by generative language models and provides empirical ground for linguistic typology research (e.g., uncovering which linguistic aspects are more universal, and which more language-specific).


Relearning lexical embeddings using our method can still be considered an expensive solution, but training costs decrease when a smaller embedding space is scaled up to the embedding space of a larger model.
In other words, approximating a good initialisation of the embedding weights decreases training time.
This method also enables adaptation of (extra) large  GPT-2 models to other languages.

If you can borrow pre-trained weights, why retrain models from scratch?
In the right column we summarise the steps for the shortest path to train your own GPT-2 for another language.


\section*{Acknowledgments}  
We gratefully acknowledge the support of the Dutch Research Council (NWO Aspasia grant for M.~Nissim) and the financial support of the Center for Groningen Language and Culture (CGTC).
Additionally, we would like to thank Lorenzo De Mattei for sharing the Italian data with us.
We would also like to thank the Center for Information Technology of the University of Groningen for providing access to the Peregrine high performance computing cluster.
Finally, we thank the anonymous reviewers for their insightful feedback.
Any mistakes remain our own.

\section*{Impact Statement} 
This work aims to minimize the environmental impact of training large neural language models by adapting existing models and by using smart initialisation of model weights.
However, experiments in this paper still require the use of GPUs for extended periods of time which has environmental impact.
Our final models are published and all models that automatically generates natural text could unfortunately be used maliciously.
While we cannot fully prevent such uses once our models are made public, we do hope that writing about risks explicitly and also raising awareness of this possibility in the general public are ways to contain the effects of potential harmful uses.
We are open to any discussion and suggestions to minimise such risks.

\vspace*{1.2cm}

\recipe{ 
This paper describes several steps that are taken to transfer GPT-2 to a different language.
The recommended shortest path to replicate this for another language is to follow these steps:

\paragraph{Vocabulary} Create a new BPE vocabulary for your target language. The optimal size for your vocabulary depends on your language, so select the size by stepwise increments until the number of tokens per sentence slows to decrease.
\paragraph{Start small} Re-initialise the lexical embeddings of the small GPT-2 model for your vocabulary size and only retrain the lexical embeddings.
\paragraph{Increase model size} If you want to train a larger model size, fit a least-squares regression model to the English lexical embeddings in the small and larger model size and use the fitted model to transform your newly trained lexical embeddings to a larger model size.
\paragraph{Optimise your embeddings} Do additional lexical embedding training in the target model size. Transformed embeddings are a good initialisation, but 
they are not perfect.
\paragraph{Finetune} Unfreeze the full target model and do some finetuning to make sure that syntax differences are learned by the new model. Use a low learning rate like $10\textsuperscript{-5}$.}{\textbf{\large Create your own GPT-2 model}}


\bibliography{paper}

\begin{thebibliography}{39}
\expandafter\ifx\csname natexlab\endcsname\relax\def\natexlab#1{#1}\fi

\bibitem[{Artetxe et~al.(2016)Artetxe, Labaka, and
  Agirre}]{artetxe_learning_2016}
Mikel Artetxe, Gorka Labaka, and Eneko Agirre. 2016.
\newblock \href {https://doi.org/10.18653/v1/D16-1250} {Learning principled
  bilingual mappings of word embeddings while preserving monolingual
  invariance}.
\newblock In \emph{Proceedings of the 2016 {Conference} on {Empirical}
  {Methods} in {Natural} {Language} {Processing}}, pages 2289--2294, Austin,
  Texas. Association for Computational Linguistics.

\bibitem[{Artetxe et~al.(2018)Artetxe, Labaka, Agirre, and
  Cho}]{artetxe_unsupervised_2018}
Mikel Artetxe, Gorka Labaka, Eneko Agirre, and Kyunghyun Cho. 2018.
\newblock \href {http://arxiv.org/abs/1710.11041} {Unsupervised {Neural}
  {Machine} {Translation}}.
\newblock \emph{arXiv:1710.11041 [cs]}.
\newblock ArXiv: 1710.11041.

\bibitem[{Artetxe et~al.(2020)Artetxe, Ruder, and
  Yogatama}]{artetxe_cross-lingual_2020}
Mikel Artetxe, Sebastian Ruder, and Dani Yogatama. 2020.
\newblock \href {https://doi.org/10.18653/v1/2020.acl-main.421} {On the
  {Cross}-lingual {Transferability} of {Monolingual} {Representations}}.
\newblock In \emph{Proceedings of the 58th {Annual} {Meeting} of the
  {Association} for {Computational} {Linguistics}}, pages 4623--4637, Online.
  Association for Computational Linguistics.

\bibitem[{Baroni et~al.(2009)Baroni, Bernardini, Ferraresi, and
  Zanchetta}]{baroni_wacky_2009}
Marco Baroni, Silvia Bernardini, Adriano Ferraresi, and Eros Zanchetta. 2009.
\newblock \href {https://doi.org/10.1007/s10579-009-9081-4} {The {WaCky} wide
  web: a collection of very large linguistically processed web-crawled
  corpora}.
\newblock \emph{Language Resources and Evaluation}, 43(3):209--226.

\bibitem[{Conneau et~al.(2020)Conneau, Khandelwal, Goyal, Chaudhary, Wenzek,
  Guzmán, Grave, Ott, Zettlemoyer, and Stoyanov}]{conneau_unsupervised_2020}
Alexis Conneau, Kartikay Khandelwal, Naman Goyal, Vishrav Chaudhary, Guillaume
  Wenzek, Francisco Guzmán, Edouard Grave, Myle Ott, Luke Zettlemoyer, and
  Veselin Stoyanov. 2020.
\newblock \href {https://doi.org/10.18653/v1/2020.acl-main.747} {Unsupervised
  {Cross}-lingual {Representation} {Learning} at {Scale}}.
\newblock In \emph{Proceedings of the 58th {Annual} {Meeting} of the
  {Association} for {Computational} {Linguistics}}, pages 8440--8451, Online.
  Association for Computational Linguistics.

\bibitem[{Conneau et~al.(2018)Conneau, Lample, Ranzato, Denoyer, and
  Jégou}]{conneau_word_2018}
Alexis Conneau, Guillaume Lample, Marc'Aurelio Ranzato, Ludovic Denoyer, and
  Hervé Jégou. 2018.
\newblock \href {http://arxiv.org/abs/1710.04087} {Word {Translation} {Without}
  {Parallel} {Data}}.
\newblock \emph{arXiv:1710.04087 [cs]}.
\newblock ArXiv: 1710.04087.

\bibitem[{De~Mattei et~al.(2020)De~Mattei, Cafagna, Dell'Orletta, Nissim, and
  Guerini}]{de_mattei_geppetto_2020}
Lorenzo De~Mattei, Michele Cafagna, Felice Dell'Orletta, Malvina Nissim, and
  Marco Guerini. 2020.
\newblock \href {http://arxiv.org/abs/2004.14253} {{GePpeTto} {Carves}
  {Italian} into a {Language} {Model}}.
\newblock \emph{arXiv:2004.14253 [cs]}.

\bibitem[{Delobelle et~al.(2020)Delobelle, Winters, and
  Berendt}]{delobelle_robbert_2020}
Pieter Delobelle, Thomas Winters, and Bettina Berendt. 2020.
\newblock \href {http://arxiv.org/abs/2001.06286} {{RobBERT}: a {Dutch}
  {RoBERTa}-based {Language} {Model}}.
\newblock \emph{arXiv:2001.06286 [cs]}.

\bibitem[{Eberhard et~al.(2020)Eberhard, Simons, and
  Fennig}]{eberhard_ethnologue_2020}
David~M. Eberhard, Gary~F. Simons, and Charles~D. Fennig. 2020.
\newblock \href {https://www.ethnologue.com} {Ethnologue: {Languages} of the
  {World}. {Twenty}-third edition.}
\newblock SIL International.

\bibitem[{Falcon(2019)}]{falcon_pytorch_2019}
W.A. Falcon. 2019.
\newblock \href {https://github.com/PyTorchLightning/pytorch-lightning}
  {{PyTorch} {Lightning}}.

\bibitem[{Glavaš et~al.(2019)Glavaš, Litschko, Ruder, and
  Vulić}]{glavas_how_2019}
Goran Glavaš, Robert Litschko, Sebastian Ruder, and Ivan Vulić. 2019.
\newblock \href {https://doi.org/10.18653/v1/P19-1070} {How to ({Properly})
  {Evaluate} {Cross}-{Lingual} {Word} {Embeddings}: {On} {Strong} {Baselines},
  {Comparative} {Analyses}, and {Some} {Misconceptions}}.
\newblock In \emph{Proceedings of the 57th {Annual} {Meeting} of the
  {Association} for {Computational} {Linguistics}}, pages 710--721, Florence,
  Italy. Association for Computational Linguistics.

\bibitem[{Joulin et~al.(2018)Joulin, Bojanowski, Mikolov, Jégou, and
  Grave}]{joulin_loss_2018}
Armand Joulin, Piotr Bojanowski, Tomas Mikolov, Hervé Jégou, and Edouard
  Grave. 2018.
\newblock \href {https://doi.org/10.18653/v1/D18-1330} {Loss in {Translation}:
  {Learning} {Bilingual} {Word} {Mapping} with a {Retrieval} {Criterion}}.
\newblock In \emph{Proceedings of the 2018 {Conference} on {Empirical}
  {Methods} in {Natural} {Language} {Processing}}, pages 2979--2984, Brussels,
  Belgium. Association for Computational Linguistics.

\bibitem[{Joulin et~al.(2017)Joulin, Grave, Bojanowski, and
  Mikolov}]{joulin_bag_2017}
Armand Joulin, Edouard Grave, Piotr Bojanowski, and Tomas Mikolov. 2017.
\newblock \href {https://www.aclweb.org/anthology/E17-2068} {Bag of {Tricks}
  for {Efficient} {Text} {Classification}}.
\newblock In \emph{Proceedings of the 15th {Conference} of the {European}
  {Chapter} of the {Association} for {Computational} {Linguistics}: {Volume} 2,
  {Short} {Papers}}, pages 427--431, Valencia, Spain. Association for
  Computational Linguistics.

\bibitem[{Kim et~al.(2019)Kim, Gao, and Ney}]{kim_effective_2019}
Yunsu Kim, Yingbo Gao, and Hermann Ney. 2019.
\newblock \href {https://doi.org/10.18653/v1/P19-1120} {Effective
  {Cross}-lingual {Transfer} of {Neural} {Machine} {Translation} {Models}
  without {Shared} {Vocabularies}}.
\newblock In \emph{Proceedings of the 57th {Annual} {Meeting} of the
  {Association} for {Computational} {Linguistics}}, pages 1246--1257, Florence,
  Italy. Association for Computational Linguistics.

\bibitem[{Kingma and Ba(2017)}]{kingma_adam_2017}
Diederik~P. Kingma and Jimmy Ba. 2017.
\newblock \href {http://arxiv.org/abs/1412.6980} {Adam: {A} {Method} for
  {Stochastic} {Optimization}}.
\newblock In \emph{{arXiv}:1412.6980 [cs]}.

\bibitem[{Kocmi and Bojar(2018)}]{kocmi_trivial_2018}
Tom Kocmi and Ondřej Bojar. 2018.
\newblock \href {https://doi.org/10.18653/v1/W18-6325} {Trivial {Transfer}
  {Learning} for {Low}-{Resource} {Neural} {Machine} {Translation}}.
\newblock In \emph{Proceedings of the {Third} {Conference} on {Machine}
  {Translation}: {Research} {Papers}}, pages 244--252, Brussels, Belgium.
  Association for Computational Linguistics.

\bibitem[{Levy et~al.(2015)Levy, Goldberg, and Dagan}]{levy_improving_2015}
Omer Levy, Yoav Goldberg, and Ido Dagan. 2015.
\newblock \href {https://doi.org/10.1162/tacl_a_00134} {Improving
  {Distributional} {Similarity} with {Lessons} {Learned} from {Word}
  {Embeddings}}.
\newblock \emph{Transactions of the Association for Computational Linguistics},
  3:211--225.

\bibitem[{Lin et~al.(2019)Lin, Chen, Lee, Li, Zhang, Xia, Rijhwani, He, Zhang,
  Ma, Anastasopoulos, Littell, and Neubig}]{lin_choosing_2019}
Yu-Hsiang Lin, Chian-Yu Chen, Jean Lee, Zirui Li, Yuyan Zhang, Mengzhou Xia,
  Shruti Rijhwani, Junxian He, Zhisong Zhang, Xuezhe Ma, Antonios
  Anastasopoulos, Patrick Littell, and Graham Neubig. 2019.
\newblock \href {https://doi.org/10.18653/v1/P19-1301} {Choosing {Transfer}
  {Languages} for {Cross}-{Lingual} {Learning}}.
\newblock In \emph{Proceedings of the 57th {Annual} {Meeting} of the
  {Association} for {Computational} {Linguistics}}, pages 3125--3135, Florence,
  Italy. Association for Computational Linguistics.

\bibitem[{Micikevicius et~al.(2018)Micikevicius, Narang, Alben, Diamos, Elsen,
  Garcia, Ginsburg, Houston, Kuchaiev, Venkatesh, and
  Wu}]{micikevicius_mixed_2018}
Paulius Micikevicius, Sharan Narang, Jonah Alben, Gregory Diamos, Erich Elsen,
  David Garcia, Boris Ginsburg, Michael Houston, Oleksii Kuchaiev, Ganesh
  Venkatesh, and Hao Wu. 2018.
\newblock \href {http://arxiv.org/abs/1710.03740} {Mixed {Precision}
  {Training}}.
\newblock \emph{arXiv:1710.03740 [cs, stat]}.
\newblock ArXiv: 1710.03740.

\bibitem[{Mikolov et~al.(2013)Mikolov, Le, and
  Sutskever}]{mikolov_exploiting_2013}
Tomas Mikolov, Quoc~V. Le, and Ilya Sutskever. 2013.
\newblock \href {http://arxiv.org/abs/1309.4168} {Exploiting {Similarities}
  among {Languages} for {Machine} {Translation}}.
\newblock \emph{arXiv:1309.4168 [cs]}.

\bibitem[{Mohiuddin et~al.(2020)Mohiuddin, Bari, and
  Joty}]{mohiuddin_lnmap_2020}
Tasnim Mohiuddin, M~Saiful Bari, and Shafiq Joty. 2020.
\newblock \href {https://www.aclweb.org/anthology/2020.emnlp-main.215}
  {{LNMap}: {Departures} from {Isomorphic} {Assumption} in {Bilingual}
  {Lexicon} {Induction} {Through} {Non}-{Linear} {Mapping} in {Latent}
  {Space}}.
\newblock In \emph{Proceedings of the 2020 {Conference} on {Empirical}
  {Methods} in {Natural} {Language} {Processing} ({EMNLP})}, pages 2712--2723,
  Online. Association for Computational Linguistics.

\bibitem[{Nguyen and Chiang(2017)}]{nguyen_transfer_2017}
Toan~Q. Nguyen and David Chiang. 2017.
\newblock \href {https://www.aclweb.org/anthology/I17-2050} {Transfer
  {Learning} across {Low}-{Resource}, {Related} {Languages} for {Neural}
  {Machine} {Translation}}.
\newblock In \emph{Proceedings of the {Eighth} {International} {Joint}
  {Conference} on {Natural} {Language} {Processing} ({Volume} 2: {Short}
  {Papers})}, pages 296--301, Taipei, Taiwan. Asian Federation of Natural
  Language Processing.

\bibitem[{Novikova et~al.(2018)Novikova, Dušek, and
  Rieser}]{novikova_rankme_2018}
Jekaterina Novikova, Ondřej Dušek, and Verena Rieser. 2018.
\newblock \href {https://doi.org/10.18653/v1/N18-2012} {{RankME}: {Reliable}
  {Human} {Ratings} for {Natural} {Language} {Generation}}.
\newblock In \emph{Proceedings of the 2018 {Conference} of the {North}
  {American} {Chapter} of the {Association} for {Computational} {Linguistics}:
  {Human} {Language} {Technologies}, {Volume} 2 ({Short} {Papers})}, pages
  72--78, New Orleans, Louisiana. Association for Computational Linguistics.

\bibitem[{Nozza et~al.(2020)Nozza, Bianchi, and Hovy}]{nozza_what_2020}
Debora Nozza, Federico Bianchi, and Dirk Hovy. 2020.
\newblock \href {http://arxiv.org/abs/2003.02912} {What the [{MASK}]? {Making}
  {Sense} of {Language}-{Specific} {BERT} {Models}}.
\newblock \emph{arXiv:2003.02912 [cs]}.

\bibitem[{Oostdijk et~al.(2013)Oostdijk, Reynaert, Hoste, and
  Schuurman}]{oostdijk_construction_2013}
Nelleke Oostdijk, Martin Reynaert, Véronique Hoste, and Ineke Schuurman. 2013.
\newblock \href {https://doi.org/10.1007/978-3-642-30910-6_13} {The
  {Construction} of a 500-{Million}-{Word} {Reference} {Corpus} of
  {Contemporary} {Written} {Dutch}}.
\newblock In Peter Spyns and Jan Odijk, editors, \emph{Essential {Speech} and
  {Language} {Technology} for {Dutch}: {Results} by the {STEVIN} programme},
  Theory and {Applications} of {Natural} {Language} {Processing}, pages
  219--247. Springer, Berlin, Heidelberg.

\bibitem[{{Ordelman, Roeland J.F.} et~al.(2007){Ordelman, Roeland J.F.}, {de
  Jong, Franciska M.G.}, {van Hessen, Adrianus J.}, and {Hondorp,
  G.H.W.}}]{ordelman_roeland_jf_twnc_2007}
{Ordelman, Roeland J.F.}, {de Jong, Franciska M.G.}, {van Hessen, Adrianus J.},
  and {Hondorp, G.H.W.} 2007.
\newblock \href
  {https://research.utwente.nl/en/publications/twnc-a-multifaceted-dutch-news-corpus(42e3c501-6cab-4212-81a9-029a774fffae).html}
  {{TwNC}: a {Multifaceted} {Dutch} {News} {Corpus}}.
\newblock \emph{ELRA Newsletter}, 12(3-4).

\bibitem[{Paszke et~al.(2019)Paszke, Gross, Massa, Lerer, Bradbury, Chanan,
  Killeen, Lin, Gimelshein, Antiga, Desmaison, Kopf, Yang, DeVito, Raison,
  Tejani, Chilamkurthy, Steiner, Fang, Bai, and Chintala}]{paszke_pytorch_2019}
Adam Paszke, Sam Gross, Francisco Massa, Adam Lerer, James Bradbury, Gregory
  Chanan, Trevor Killeen, Zeming Lin, Natalia Gimelshein, Luca Antiga, Alban
  Desmaison, Andreas Kopf, Edward Yang, Zachary DeVito, Martin Raison, Alykhan
  Tejani, Sasank Chilamkurthy, Benoit Steiner, Lu~Fang, Junjie Bai, and Soumith
  Chintala. 2019.
\newblock \href
  {http://papers.nips.cc/paper/9015-pytorch-an-imperative-style-high-performance-deep-learning-library.pdf}
  {{PyTorch}: {An} {Imperative} {Style}, {High}-{Performance} {Deep} {Learning}
  {Library}}.
\newblock In H.~Wallach, H.~Larochelle, A.~Beygelzimer,
  F.~d{\textbackslash}textquotesingle Alché-Buc, E.~Fox, and R.~Garnett,
  editors, \emph{Advances in {Neural} {Information} {Processing} {Systems} 32},
  pages 8026--8037. Curran Associates, Inc.

\bibitem[{Peirce et~al.(2019)Peirce, Gray, Simpson, MacAskill, Höchenberger,
  Sogo, Kastman, and Lindeløv}]{peirce_psychopy2_2019}
Jonathan Peirce, Jeremy~R. Gray, Sol Simpson, Michael MacAskill, Richard
  Höchenberger, Hiroyuki Sogo, Erik Kastman, and Jonas~Kristoffer Lindeløv.
  2019.
\newblock \href {https://doi.org/10.3758/s13428-018-01193-y} {{PsychoPy2}:
  {Experiments} in behavior made easy}.
\newblock \emph{Behavior Research Methods}, 51(1):195--203.

\bibitem[{Radford et~al.(2019)Radford, Wu, Child, Luan, Amodei, and
  Sutskever}]{radford_language_2019}
Alec Radford, Jeffrey Wu, Rewon Child, David Luan, Dario Amodei, and Ilya
  Sutskever. 2019.
\newblock \href {https://openai.com/blog/better-language-models/} {Language
  {Models} are {Unsupervised} {Multitask} {Learners}}.

\bibitem[{Ruder et~al.(2019)Ruder, Vulić, and Søgaard}]{ruder_survey_2019}
Sebastian Ruder, Ivan Vulić, and Anders Søgaard. 2019.
\newblock \href {https://doi.org/10.1613/jair.1.11640} {A {Survey} of
  {Cross}-lingual {Word} {Embedding} {Models}}.
\newblock \emph{Journal of Artificial Intelligence Research}, 65:569--631.

\bibitem[{Smith(2017)}]{smith_cyclical_2017}
Leslie~N. Smith. 2017.
\newblock \href {https://doi.org/10.1109/WACV.2017.58} {Cyclical {Learning}
  {Rates} for {Training} {Neural} {Networks}}.
\newblock In \emph{2017 {IEEE} {Winter} {Conference} on {Applications} of
  {Computer} {Vision} ({WACV})}, pages 464--472.

\bibitem[{Strubell et~al.(2019)Strubell, Ganesh, and
  McCallum}]{strubell_energy_2019}
Emma Strubell, Ananya Ganesh, and Andrew McCallum. 2019.
\newblock \href {https://doi.org/10.18653/v1/P19-1355} {Energy and {Policy}
  {Considerations} for {Deep} {Learning} in {NLP}}.
\newblock In \emph{Proceedings of the 57th {Annual} {Meeting} of the
  {Association} for {Computational} {Linguistics}}, pages 3645--3650, Florence,
  Italy. Association for Computational Linguistics.

\bibitem[{Søgaard et~al.(2018)Søgaard, Ruder, and
  Vulić}]{sogaard_limitations_2018}
Anders Søgaard, Sebastian Ruder, and Ivan Vulić. 2018.
\newblock \href {https://doi.org/10.18653/v1/P18-1072} {On the {Limitations} of
  {Unsupervised} {Bilingual} {Dictionary} {Induction}}.
\newblock In \emph{Proceedings of the 56th {Annual} {Meeting} of the
  {Association} for {Computational} {Linguistics} ({Volume} 1: {Long}
  {Papers})}, pages 778--788, Melbourne, Australia. Association for
  Computational Linguistics.

\bibitem[{de~Vries et~al.(2020)de~Vries, van Cranenburgh, and
  Nissim}]{de_vries_whats_2020}
Wietse de~Vries, Andreas van Cranenburgh, and Malvina Nissim. 2020.
\newblock \href {https://www.aclweb.org/anthology/2020.findings-emnlp.389}
  {What's so special about {BERT}'s layers? {A} closer look at the {NLP}
  pipeline in monolingual and multilingual models}.
\newblock In \emph{Findings of the {Association} for {Computational}
  {Linguistics}: {EMNLP} 2020}, pages 4339--4350, Online. Association for
  Computational Linguistics.

\bibitem[{de~Vries et~al.(2019)de~Vries, Cranenburgh, Bisazza, Caselli, Noord,
  and Nissim}]{de_vries_bertje_2019}
Wietse de~Vries, Andreas~van Cranenburgh, Arianna Bisazza, Tommaso Caselli,
  Gertjan~van Noord, and Malvina Nissim. 2019.
\newblock \href {http://arxiv.org/abs/1912.09582} {{BERTje}: {A} {Dutch} {BERT}
  {Model}}.
\newblock \emph{arXiv:1912.09582 [cs]}.

\bibitem[{Wolf et~al.(2020)Wolf, Debut, Sanh, Chaumond, Delangue, Moi, Cistac,
  Rault, Louf, Funtowicz, and Brew}]{wolf_huggingfaces_2020}
Thomas Wolf, Lysandre Debut, Victor Sanh, Julien Chaumond, Clement Delangue,
  Anthony Moi, Pierric Cistac, Tim Rault, Rémi Louf, Morgan Funtowicz, and
  Jamie Brew. 2020.
\newblock \href {http://arxiv.org/abs/1910.03771} {{HuggingFace}'s
  {Transformers}: {State}-of-the-art {Natural} {Language} {Processing}}.
\newblock \emph{arXiv:1910.03771 [cs]}.

\bibitem[{Wu and Dredze(2020)}]{wu_are_2020}
Shijie Wu and Mark Dredze. 2020.
\newblock \href {http://arxiv.org/abs/2005.09093} {Are {All} {Languages}
  {Created} {Equal} in {Multilingual} {BERT}?}
\newblock \emph{arXiv:2005.09093 [cs]}.
\newblock ArXiv: 2005.09093.

\bibitem[{Xing et~al.(2015)Xing, Wang, Liu, and Lin}]{xing_normalized_2015}
Chao Xing, Dong Wang, Chao Liu, and Yiye Lin. 2015.
\newblock \href {https://doi.org/10.3115/v1/N15-1104} {Normalized {Word}
  {Embedding} and {Orthogonal} {Transform} for {Bilingual} {Word}
  {Translation}}.
\newblock In \emph{Proceedings of the 2015 {Conference} of the {North}
  {American} {Chapter} of the {Association} for {Computational} {Linguistics}:
  {Human} {Language} {Technologies}}, pages 1006--1011, Denver, Colorado.
  Association for Computational Linguistics.

\bibitem[{Zoph et~al.(2016)Zoph, Yuret, May, and Knight}]{zoph_transfer_2016}
Barret Zoph, Deniz Yuret, Jonathan May, and Kevin Knight. 2016.
\newblock \href {https://doi.org/10.18653/v1/D16-1163} {Transfer {Learning} for
  {Low}-{Resource} {Neural} {Machine} {Translation}}.
\newblock In \emph{Proceedings of the 2016 {Conference} on {Empirical}
  {Methods} in {Natural} {Language} {Processing}}, pages 1568--1575, Austin,
  Texas. Association for Computational Linguistics.

\end{thebibliography}
\bibliographystyle{acl_natbib}


\end{document}